\documentclass[letterpaper]{article} % DO NOT CHANGE THIS
\usepackage{aaai24}  % DO NOT CHANGE THIS
\usepackage{times}  % DO NOT CHANGE THIS
\usepackage{helvet}  % DO NOT CHANGE THIS
\usepackage{courier}  % DO NOT CHANGE THIS
\usepackage[hyphens]{url}  % DO NOT CHANGE THIS
\usepackage{graphicx} % DO NOT CHANGE THIS
\usepackage{natbib}  % DO NOT CHANGE THIS AND DO NOT ADD ANY OPTIONS TO IT
\usepackage{caption} % DO NOT CHANGE THIS AND DO NOT ADD ANY OPTIONS TO IT
\usepackage{algorithm}
\usepackage{algorithmic}

\usepackage{epsfig}
\usepackage{amsmath}
\usepackage{amssymb}

\usepackage{booktabs}
\usepackage{multirow}
\usepackage{paralist}

\usepackage{newfloat}
\usepackage{listings}
\DeclareCaptionStyle{ruled}{labelfont=normalfont,labelsep=colon,strut=off} % DO NOT CHANGE THIS
\floatstyle{ruled}
\newfloat{listing}{tb}{lst}{}
\floatname{listing}{Listing}
\title{Deep Homography Estimation for Visual Place Recognition}
\author{
    Feng Lu\textsuperscript{\rm 1,\rm 2}, Shuting Dong\textsuperscript{\rm 1,\rm 2}, Lijun Zhang\textsuperscript{\rm 3}, Bingxi Liu\textsuperscript{\rm 2,\rm4}, Xiangyuan Lan\textsuperscript{\rm 2*}, Dongmei Jiang\textsuperscript{\rm 2},\\ Chun Yuan\textsuperscript{\rm 1,\rm 2}\thanks{Corresponding authors.}
}
\affiliations{
    \textsuperscript{\rm 1}Tsinghua Shenzhen International Graduate School, Tsinghua University \quad
    \textsuperscript{\rm 2}Peng Cheng Laboratory\\
    \textsuperscript{\rm 3}Chongqing Institute of Green and Intelligent Technology, Chinese Academy of Sciences\\
    \textsuperscript{\rm 4}Southern University of Science and Technology
    \{lf22@mails, dst21@mails, yuanc@sz\}.tsinghua.edu.cn,  zhanglijun@cigit.ac.cn, \{liubx, lanxy, jiangdm\}@pcl.ac.cn 
}

\usepackage{bibentry}
\begin{document}

\maketitle

\begin{abstract}
Visual place recognition (VPR) is a fundamental task for many applications such as robot localization and augmented reality. Recently, the hierarchical VPR methods have received considerable attention due to the trade-off between accuracy and efficiency. They usually first use global features to retrieve the candidate images, then verify the spatial consistency of matched local features for re-ranking. However, the latter typically relies on the RANSAC algorithm for fitting homography, which is time-consuming and non-differentiable. This makes existing methods compromise to train the network only in global feature extraction. Here, we propose a transformer-based deep homography estimation (DHE) network that takes the dense feature map extracted by a backbone network as input and fits homography for fast and learnable geometric verification. Moreover, we design a re-projection error of inliers loss to train the DHE network without additional homography labels, which can also be jointly trained with the backbone network to help it extract the features that are more suitable for local matching. Extensive experiments on benchmark datasets show that our method can outperform several state-of-the-art methods. And it is more than one order of magnitude faster than the mainstream hierarchical VPR methods using RANSAC. The code is released at https://github.com/Lu-Feng/DHE-VPR.
\end{abstract}

\section{Introduction}

Visual place recognition (VPR), also known as visual geo-localization \cite{benchmark} or image localization \cite{liu2019}, is one of the research hotspots in robotics and computer vision communities. VPR aims to coarsely estimate the location of the query image (i.e. the current location of the mobile robot), which is commonly achieved using image retrieval methods on a database of geo-tagged images. When designing a robust VPR method, there are two challenging problems to consider:  1) Due to condition (e.g., light, weather, and season) and viewpoint variations, images captured at the same place may change significantly over time. 2) Images captured at different places can be similar, which may lead to perceptual aliasing \cite{survey}.

The VPR implementation process typically involves image retrieval and feature matching \cite{netvlad,delg}, with global or/and local features to represent place images. The global features can also be got by aggregating local features into compact feature vectors \cite{vLAD}, which are robust to viewpoint change and applicable for large-scale VPR. However, methods only using such features are prone to suffer from perceptual aliasing, because it neglects the spatial information of aggregated local features. This issue can be solved by matching local features with geometric verification, but it is time-consuming. A compromise pipeline \cite{patchvlad}, called hierarchical or two-stage VPR, first retrieves top-k candidate places using global features, then re-ranks them via local feature matching between the query and candidate images.

\begin{figure}[!t]
	\centering
	\includegraphics[width=0.94\linewidth]{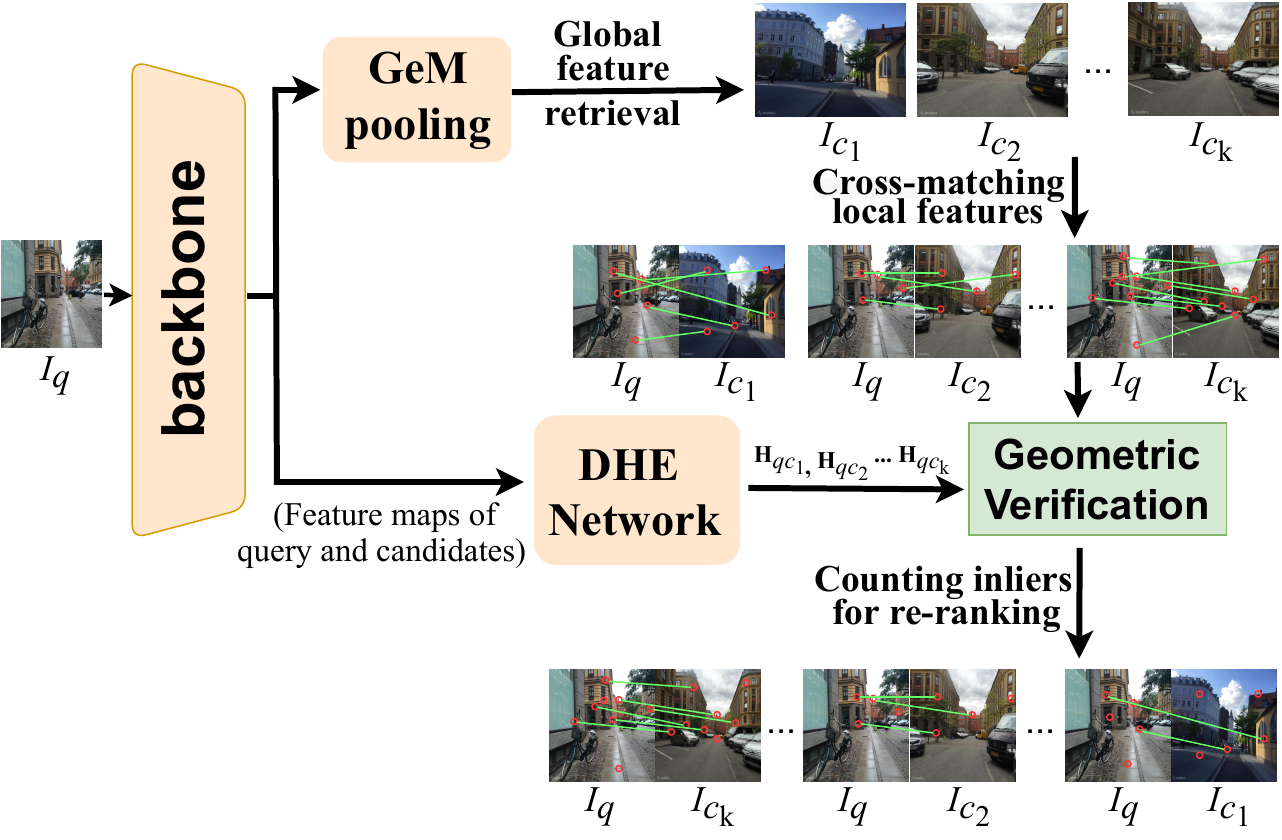}
    \vspace{-0.2cm}
	\caption{The two-stage place retrieval with the proposed architecture. The backbone is applied to extract feature maps. The top branch yields global features for retrieving top-k candidate images. The bottom branch employs the local features for cross-matching and the DHE network for geometric verification via regressing homography. We count inliers as image similarity for re-ranking candidates.
	}
    \label{architecture}
	\vspace{-0.5cm}
\end{figure}

However, in the re-ranking stage of hierarchical VPR, the matched local feature pairs based on similarity searching commonly contain some incorrect pairs (i.e., outliers). The RANSAC algorithm \cite{ransac} is typically used to fit the homography and remove outliers. The process of RANSAC is to iteratively sample four matched pairs at random to solve corresponding homography transformations and find the one that meets the most inliers. It is time-consuming and non-differentiable. Although there are some differentiable RANSAC models \cite{dsac,neuralRANSAC,dsacstar}, they are inapplicable to the VPR task. This leads to some state-of-the-art (SOTA) VPR methods \cite{patchvlad,transvpr} compromising to train the network only in global feature extraction. Deep homography estimation (\textbf{DHE}) \cite{nguyen}, which uses deep neural networks to fit homography, is a more viable solution. GeoWarp \cite{geowarp} first used a similar way in VPR to align two different views of a same place in the urban scene. Regardless of the existing deep homography methods or the GeoWarp work, they all regress a homography to align two images of a planar scene (or approximated as a plane). However, many scenes in VPR (e.g., suburban scene) do not satisfy this condition. In fact, fitting appropriate homography to remove outliers and use the number of inliers as image similarity, as in SOTA two-stage VPR works \cite{patchvlad,transvpr}, is suitable for more scenes in VPR and can achieve better performance.

In this paper, we propose a transformer-based DHE network that takes the dense feature map extracted by a backbone network as input and regresses a homography matrix to decide the inliers of matched local feature pairs as image similarity (see Fig. \ref{architecture}). Given that none of the VPR training datasets have homography annotations, we propose a re-projection error of inliers (REI) loss to train the DHE network using the reference number of inliers provided by RANSAC as supervision. This process can be jointly trained with the backbone, making the feature map extracted by the backbone more suitable for local feature matching. Since our method uses the DHE network (without RANSAC) in inference, it can also make the re-ranking stage much faster.

We name our proposed method DHE-VPR. The main \textbf{contributions} can be highlighted as follows:

\textbf{1)} We introduce a novel hierarchical VPR architecture, in which a DHE network (rather than RANSAC) is adopted to fit homography to count inliers of matched local feature pairs. This makes up for the time-consuming and non-differentiable defect of RANSAC.

\textbf{2)} We propose a re-projection error of inliers loss to train the DHE network without additional homography labels. The error can be back-propagated to the backbone, making the learned local features more suitable for re-ranking.

\textbf{3)} Extensive experiments show that the proposed method outperforms several SOTA methods. And it is more than one order of magnitude faster than the existing two-stage VPR methods with RANSAC-based geometric verification.

\section{Related Work}
\textbf{One-Stage VPR:} The early VPR methods commonly got the most similar place images through direct retrieval without considering re-ranking. We call them one-stage VPR. These methods represented the place image with global descriptors produced by aggregating local descriptors or processing the whole image. For example, aggregation algorithms like Bag of Words \cite{BoW} and VLAD \cite{vLAD,VLAD1,VLAD2} have been employed to aggregate local descriptors such as SURF \cite{SURF}. With the significant advancements of deep learning in computer vision tasks, many VPR methods \cite{landmarks, categorization1, categorization2, SMM1, netvlad, densernet, SPED, yin2019, semantic, gsv, cosplace} have opted to use deep features for image representation to achieve better performance. Similarly, several works \cite{netvlad,DBOW,attentionVLAD,speNetvlad} also incorporated the traditional aggregation models into neural networks. However, relying solely on aggregated features often leads to perceptual aliasing due to the lack of spatial information. To overcome this, some approaches utilized image sequence matching \cite{seqslam,hmmpr2,naseer, hmmpr,sta-vpr,seqnet} to achieve robust VPR under extreme variations in illumination, weather, and season. And other methods \cite{landmarks,landmarks2,landmarks3,gao2020} mined discriminative landmarks for VPR. Moreover, deep neural networks commonly require training on large-scale place datasets with weak geographic supervision. CRN \cite{crn} and SFRS \cite{sfrs} mined hard positive samples instead of the simplest positive sample \cite{netvlad} for training more robust VPR networks.

\textbf{Two-Stage VPR:} More recently, the hierarchical strategy with re-ranking candidate images for VPR \cite{hvpr,patchvlad,geowarp,transvpr,aanet,etr} has gained more attention. The hierarchical (two-stage) VPR methods commonly first searched top-k candidate places using compact global descriptors, and then re-ranked candidates by local feature matching. The global features are typically produced using aggregation/pooling methods, e.g., NetVLAD \cite{netvlad} and Generalized Mean (GeM) pooling \cite{gem}. The re-ranking stage usually involves geometric verification \cite{patchvlad,delg,transvpr}, which takes into account the spatial relationship of local features and thus is able to address the perceptual aliasing encountered by global features. However, most of them required the RANSAC algorithm \cite{ransac} for fitting homography, which is time-consuming and non-differentiable. GeoWarp \cite{geowarp} first applied a convolutional neural network to regress a homography transformation for aligning two images taken at a same place with different viewpoints. However, the aligned two images need to be in a plane, which can be satisfied mainly in the urban environment. Our work can be seen as a method that draws on the advantages of both geometric validation and deep homography to achieve better performance and higher efficiency, and is also applicable to various scenarios in VPR.

\textbf{Deep Homography Estimation:} Homography estimation is a basic problem in computer vision. Most traditional methods used Direct Linear Transform \cite{dlt} with RANSAC outlier rejection for homography estimation. In recent years, there has been a surge of interest in developing deep neural networks for this task. DeTone et al. \cite{DeTone} designed a VGG-style network to estimate homography and demonstrated its effectiveness. Nguyen et al. \cite{nguyen} proposed an unsupervised approach that optimizes the DHE network by minimizing the pixel-wise intensity error between a warped input image and the other image. Further, Zhang et al. \cite{zhang} began to calculate loss (i.e. image distance) in feature space instead of pixel space. Koguciuk et al. \cite{koguciuk} presented a bidirectional implicit homography estimation loss for unsupervised training. Le et al. \cite{le} developed a model that can jointly estimate the homography and dynamics masks to handle dynamic scenes. However, all these works are fitting homography to align images. To the best of our knowledge, our work is among the first to use neural networks to fit homography for geometric check. The homography matrices used for alignment and geometric verification are not necessarily the same. And the loss in our method is based on the re-projection error, which is different from the loss based on image distance in the above methods.

\section{Methodology}
%This section introduces the proposed method in detail. We begin by outlining the overall architecture, as well as describing how to retrieve candidate places and re-rank the candidates. Then we detail the process of using the DHE network to get the required homography matrix. Following that, we use the homography to perform geometric verification on matched local feature pairs and calculate image similarity scores. Finally, we present our proposed training strategy for training the entire model.

\subsection{Problem Formulation}
Given a query image $I_q$ of a previously visited place, the task of a VPR system is to find its best match from a database of geo-tagged place images $\mathcal{D} = \{I_i\}$. The two-stage VPR methods typically first perform a similarity retrieval over $\mathcal{D}$ in the space of the global features to yield a set of candidate images $\mathcal{C} = \{I_c\}$ $(\mathcal{C} \subset \mathcal{D})$, i.e. search top-k candidates. Then, the local matching (and geometric verification) algorithms are used to re-rank the candidate images in $\mathcal{C}$ based on local features. Specifically, we use a neural network as unified feature extractor to get the feature map $\boldsymbol{f} \in \mathbb{R}^{W \times H \times C}$ ($weight \times height \times channel$) of each place image. In the first stage, $\boldsymbol{f}$ is aggregated/pooled into a compact vector as global feature. In the second stage, $\boldsymbol{f}$ is directly viewed as a dense $W \times H$ grid of $C$-dimensional local features for re-ranking. The local features are L2-normalized.

\subsection{Architecture Overview}
Due to the superiority of Vision Transformer \cite{vit} in capturing feature dependencies over long distances, the Compact Convolutional Transformer (CCT) \cite{cct} is used as the unified feature extractor (i.e. backbone) in this work. Its output is a $M\times C$-dimensional tensor ($M$ means the number of patch tokens), which can be reshaped into the feature map $\boldsymbol{f} \in \mathbb{R}^{W \times H \times C} (W\times H = M$) to restore spatial position. As shown in Fig. \ref{architecture}, the proposed architecture consists of two branches. In the above branch, GeM pooling \cite{gem} is utilized to aggregate the feature map into $C$-dimensional vector, i.e., global feature. Then L2 distance is used to measure the global feature distance between the query image $I_q$ and each reference image $I_i$ in $\mathcal{D}$ to get candidate image set $\mathcal{C}$. The bottom branch is primarily composed of our proposed DHE network, the inputs of which are the feature maps (i.e. dense local features) of the query and candidate images. It uses the homography estimated by the DHE network to check the geometric consistency of matched local feature pairs between the query $I_q$ and each candidate $I_c$ in $\mathcal{C}$. The number of inliers is used as the similarity of image pairs to re-rank candidate images. 

\subsection{Deep Homography Estimation Network}
\begin{figure*}[!t]
	\centering
	\includegraphics[width=0.88\linewidth]{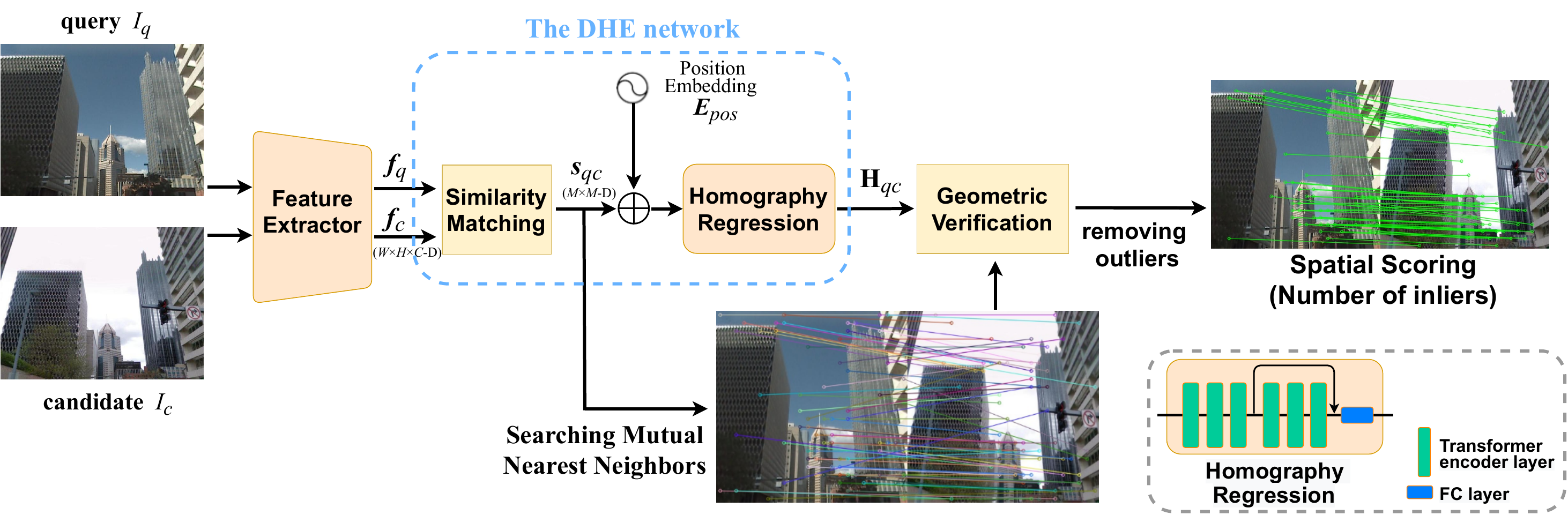}
  \vspace{-0.2cm}
	\caption{
		Diagram of our re-ranking process with the DHE network. The feature maps $\boldsymbol{f}_q$ and $\boldsymbol{f}_c$ of the query image $I_q$ and a candidate image $I_c$ are fed into the Similarity Matching Module to compute the similarity map $\boldsymbol{s}_{qc}$. Then the Homography Regression Module uses the $\boldsymbol{s}_{qc}$ to yield the homography matrix $\mathbf{H}_{qc}$ for geometric verification of mutual matches.
	}
	\label{DHE}
 \vspace{-0.4cm}
\end{figure*}

Based on the multi-view geometry theory, we can use a homography matrix to associate two images presenting a same planar scene or captured by a rotational camera. When homogeneous coordinates are used to denote points, the point $(u, v)^T$ can be represented as $(u, v, 1)^T$. Meanwhile, the homogeneous coordinates $(x, y, z)^T$ and $(x/z, y/z, 1)^T$ denote a same point. Given two points $\mathbf{x} = (u, v, 1)^T$ and $\mathbf{x}' = (u', v', 1)^T$, we can use a non-singular $3 \times 3$ matrix $\mathbf{H}$ to express the homography transformation mapping $\mathbf{x} \leftrightarrow \mathbf{x}'$:

\begin{align}
    \begin{bmatrix}
    u' \\ 
    v'   \\ 
    1  
    \end{bmatrix} 
  & \simeq \begin{bmatrix}
    h_{11} & h_{12} & h_{13} \\ 
    h_{21} & h_{22} & h_{23} \\ 
    h_{31} & h_{32} & h_{33} 
    \end{bmatrix} 
    \begin{bmatrix}
    u \\ 
    v  \\ 
    1  
    \end{bmatrix} \hspace{.1in} \mbox{   or  } \hspace{.1in} \mathbf{x}' \simeq \mathbf{H}\mathbf{x}.
\label{eq:H}
\end{align}

Even when multiplied by any non-zero scale factor, $\mathbf{H}$ does not change the projective transformation, and only the ratio of matrix elements is meaningful. We can set $h_{33}=1$ and keep 8 independent ratios in $\mathbf{H}$ as 8 degrees of freedom. So the homography matrix $\mathbf{H}$ can be solved using 4 pairs (non-collinear) of corresponding points in two images.

Previous deep homography works utilize deep neural networks to output the homography for warping an image to another. In VPR, the building facades in urban environments are typically planes, so GeoWarp \cite{geowarp} employed deep homography to warp two place images taken from different viewpoints of the same place to a closer geometrical space. However, in more general scenes, the content of the whole image or a large region of the image is not exactly in a plane, so aligning two place images with homography is ineffective. Here, we regress a homography matrix to check spatial consistency instead of aligning images. The purpose of this work is different from previous deep homography works. However, we use a network to regress homography, which can learn from previous works.

As shown in Fig. \ref{DHE}, feature maps $\boldsymbol{f}_q$ and $\boldsymbol{f}_c$ of the query image $I_q$ and a candidate image $I_c$ are first extracted using feature extractor (i.e. the backbone), and then are fed to our DHE network. The DHE network consists of two modules: Similarity Matching Module and Homography Regression Module. The former is a parameterless and differentiable operation. The latter is a learnable neural network.

The Similarity Matching Module $S$ utilizes the cosine similarity to compute the similarity map $\boldsymbol{s}_{qc} \in \mathbb{R}^{M \times M}$ between local features pairs of $\boldsymbol{f}_q$ and $\boldsymbol{f}_c$. The $M$ is the number of local features (i.e. patch tokens for Transformer) in each image. The $\boldsymbol{s}_{qc}$ can also be expressed as $S(\boldsymbol{f}_q, \boldsymbol{f}_c)$. Since the local features have been L2-normalized, the cosine similarity can be replaced by the inner product. Formally,
\begin{equation}
   \boldsymbol{s}_{qc}(i,j) = \boldsymbol{f}_q(i)^\text{T} \boldsymbol{f}_c(j)  \quad i,j \in \{1,2,...,M\}.
\end{equation}

The Homography Regression Module applies the similarity map $\boldsymbol{s}_{qc}$ plus a learnable position embedding $\boldsymbol{E}_{pos}\in \mathbb{R}^{M \times M}$ as input and yields the homography matrix $\mathbf{H}_{qc}$ for following geometric verification. However, it is difficult to directly regress the elements in $\mathbf{H}$ due to the high variance in their magnitude. As a result, it is common to use networks to regress the 4 pairs of corresponding points (or 4-point offsets) in two images \cite{DeTone,nguyen,geowarp}. Inspired by this, this module first computes the 4 pairs of points (4-point correspondence) on the images $I_q$ and $I_c$, i.e., a 16D vector. This process is implemented by using 6 stacked transformer encoder layers followed by a fully connected layer, and a shortcut connection is used between the third encoder layer and the sixth encoder layer. The transformer encoder can provide a larger effective perceptual field on the similarity map. And it is denoted as
\begin{equation}
    \label{eq:regression}
    \begin{aligned}
    TransE(\boldsymbol{s}_{qc}+\boldsymbol{E}_{pos})&=[\underbrace{\boldsymbol{p}_{q1}, \ldots, \boldsymbol{p}_{q4}}_{\boldsymbol{P}_q}, \underbrace{\boldsymbol{p}_{c1}, \ldots, \boldsymbol{p}_{c4}}_{\boldsymbol{P}_c}] \\
    &= [\boldsymbol{P}_q, \boldsymbol{P}_c],
    \end{aligned}
\end{equation}
where $\boldsymbol{p}_{q1}, \ldots, \boldsymbol{p}_{q4}$ and $\boldsymbol{p}_{c1}, \ldots, \boldsymbol{p}_{c4}$ are four points on $I_q$ and $I_c$, respectively. $[\boldsymbol{P}_q, \boldsymbol{P}_c]$ is their concatenation.

After obtaining 4 pairs of points on two images, we use the Direct Linear Transform \cite{dlt} to yield the homography $\mathbf{H}_{qc}$, which is represented as
\begin{equation}
    \mathbf{H}_{qc} = DLT(\mathbf{P}_q,\mathbf{P}_c).
\end{equation}
We succinctly denote the whole process of this module as
\begin{equation}
    \mathbf{H}_{qc} = R(\boldsymbol{s}_{qc}).
\end{equation}
The whole process of the DHE network is summarized as
\begin{equation}  
\label{eq:warping_detailed}
    \mathbf{H}_{qc} = DHE(\boldsymbol{f}_q, \boldsymbol{f}_c) = R\big( S(\boldsymbol{f}_q, \boldsymbol{f}_c) \big).
\end{equation}

\subsection{Re-ranking with Geometric Verification}
In the re-ranking stage, we first calculate the mutual nearest neighbor matching of local features through exhaustive cross-matching. Then the homography matrix yielded by the DHE network is used to check geometric verification and remove outliers. Finally, we use the number of inliers as the similarities of image pairs to re-rank the candidates.

The similarity map $\boldsymbol{s}_{qc}$ between the query $I_q$ and a candidate $I_c$ has been computed. We apply it to search mutual nearest neighbor matches set $\mathcal{MN}$, which is defined as 

\begin{equation}
\label{eq::mn}
    \begin{aligned}
        \mathcal{MN} = \{\left(x, y\right): \; 
        x & = \mathop{\arg\max}_{i}\boldsymbol{s}_{qc}(i,y),\\
        y & = \mathop{\arg\max}_{j}\boldsymbol{s}_{qc}(x,j)\}.
    \end{aligned}
\end{equation}
That is, if the most similar local feature in image $I_q$ of the feature $\boldsymbol{f}_c(y)$ is $\boldsymbol{f}_q(x)$, and the most similar feature in image $I_c$ of the feature $\boldsymbol{f}_q(x)$ is $\boldsymbol{f}_c(y)$, then $(x,y) \in \mathcal{MN}$.

The $x$ and $y$ are the indices of corresponding patches. Assume that each patch corresponds to the Cartesian coordinates of a 2D image point in the center of the patch. We denote the homogeneous coordinates of a pair of matched patches as $\mathbf{p}_q$ and $\mathbf{p}_c$. If the re-projection error is less than a threshold $\theta$, then it is an inlier. That is, an inlier satisfies
\begin{equation}
       \|C(\mathbf{p}_c)-C(\mathbf{H}_{qc}\mathbf{p}_q)\| \leq \theta,
\end{equation}
where $C(\cdot)$ means converting homogeneous coordinates to Cartesian (inhomogeneous) coordinates. We re-rank the candidate images using the number of inliers between the query and candidates.

\subsection{Training Strategy} 
In common deep homography works, an image can be aligned with another image after using the homography transformation because the two images present a (approximately) same planar scene. So, they can train networks by minimizing the distance between two aligned images. But this prior condition does not hold for place images in many scenes. Meanwhile, we also cannot directly obtain homography labels to supervise the training of our DHE network.

One possible solution is to train the network with the label provided by RANSAC. However, we found that the network is difficult to converge whether using the homography parameters yielded by RANSAC for direct supervision, or using the selected inliers and their re-projection errors for supervision. We propose to use only the number of inliers provided by RANSAC as supervision information, let the network autonomously decide which matched pairs are inliers, and optimize the network by minimizing the average re-projection error of these ``inliers". This makes it easier for networks to converge. Meanwhile, the loss produced by this process has the potential to optimize the feature extractor through back-propagation, so that the extracted feature maps are more suitable for local matching and outperform the method directly using RANSAC.

We first initialize the backbone and the DHE network individually. The initialization of the backbone follows the training procedure of NetVLAD using triplet loss. That is
\begin{equation}
L_g=\sum_{j} l\left(d_{G}\left(q, p^{q}\right)+m-d_{G}\left(q, n_{j}^{q}\right)\right),
\label{eq:hinge}
\end{equation}
where $l(x) = \max(x, 0)$. $m$ is the margin. $d_G$ is the L2 distance between global features of two images. $q$, $p^{q}$, and $n_{j}^q$ are the query, positive sample, and hard negative sample.
 
The feature maps extracted from the trained backbone are fed to the DHE network to yield homography $\mathbf{H}$. Meanwhile, we use Eq. \ref{eq::mn} to compute the mutual matches of local features in the positive image pair, and apply RANSAC to decide the number of inliers $N$. For all mutual matches, $\mathbf{H}$ is used for transformation. $N$ pairs of matched points with the smallest re-projection error are regarded as ``inliers". The set of the homogeneous coordinates of these ``inliers" is denoted as $\mathcal{IN}$. We aim to minimize their re-projection errors. So we design a re-projection error of inliers (REI) loss for the DHE network training. That is
\begin{equation}
L_r = \frac{\sum_{(\mathbf{\hat{p}}_c,\mathbf{\hat{p}}_q) \in \mathcal{IN}}\|C(\mathbf{\hat{p}}_c)-C(\mathbf{H}_{qc}\mathbf{\hat{p}}_q)\|}{N}.
\end{equation}

After the initialization of the backbone and DHE network, they are fine-tuned together using the joint loss:
\begin{equation}
 L = L_g + \lambda L_r,
 \label{eq:jloss}
\end{equation}
where $\lambda$ is a weight. Note that when we initialize the DHE network, the backbone is frozen. But when we combine them for fine-tuning, the parameters of the last few layers of the backbone are updatable.

\section{Experiments}
%In this section, we first describe the datasets, evaluation metrics, and implementation details. Then we validate the design of our proposed architecture using ablation experiments. Finally, we compare our approach with several SOTA methods and show its advantage in computational efficiency.

\subsection{Datasets and Performance Evaluation}
We conduct experiments using multiple VPR datasets: \textbf{MSLS}  \cite{msls}, \textbf{Pitts30k} \cite{pitts}, \textbf{Nordland} (downsampled test set with 224x224 image size) \cite{nordland}, and \textbf{St. Lucia} \cite{benchmark}. Table \ref{tableDataset} summarizes their main information. The Recall@N (R@N) is used to assess model performance, which calculates the percentage of queries that have at least one of the top-N retrieved reference images taken within a certain threshold of ground truth. Following common procedure \cite{transvpr,benchmark}, the threshold is 25m and 40$^{\circ}$ for MSLS (including MSLS val and MSLS challenge), 25m for Pitts30k and St. Lucia, and $\pm 2$ frames for Nordland. 

\begin{table}[!t]
\begin{center}
\small
  \setlength{\tabcolsep}{1.0mm}{
  \renewcommand{\arraystretch}{0.97}
  \begin{tabular}{cccc}
    \toprule
    \multirow{2}{*}{Dataset} & \multicolumn{1}{c}{\multirow{2}{*}{Description}} &  \multicolumn{2}{c}{Variation} \\
     \multicolumn{1}{c}{}& & \multicolumn{1}{c}{Condition} & \multicolumn{1}{c}{Viewpoint} \\ 
    \hline
    MSLS & long-term, urban, suburban  &  \checkmark & \checkmark \\
 \hline
    Pitts30k & urban, panorama  & \checkmark & \checkmark \\
 \hline
    Nordland & suburban, natural, seasonal & \checkmark & $\times$ \\ 
 \hline
    St. Lucia & suburban  & \checkmark & \checkmark \\
    \bottomrule
  \end{tabular}}
\end{center}
\vspace{-0.2cm}
\caption{Summary of the datasets.}
\vspace{-0.4cm}
\label{tableDataset}
\end{table}

\subsection{Implementation Details}
In DHE-VPR architecture, the CCT-14 model pre-trained on ImageNet \cite{ImageNet} is used as the backbone. The transformer encoder layers after the 8th layer are removed, before the 3rd layer are frozen. We resize the input image to 384×384 pixels and get 24×24×384-D feature maps (the global features are 384-D). We re-rank the top-32 candidates to yield final results. The re-projection error threshold $\theta$ of the inlier is set to 1.5 times the patch size for RANSAC, and 3 times the patch size for geometric verification using DHE (in inference). The margin $m$ in Eq. \ref{eq:hinge} is set to 0.1, and the weight $\lambda$ in Eq. \ref{eq:jloss} is 100. Experiments are implemented using PyTorch on an NVIDIA GeForce RTX 3090 GPU. 
For the initialization of the DHE network, the Adam optimizer is used with learning rate = 0.0001 (multiplied by 0.8 after every 5 epochs) and batch size = 16. We train the network for 100 epochs (2k iterations per epoch) on MSLS-train. The implementation of the backbone initialization and the fine-tuning of entire network basically follows the benchmark \cite{benchmark}, with learning rate = 0.00001 and batch size = 4. For the backbone initialization, we train CCT-14 on MSLS-train for MSLS, Nordland, and St. Lucia, and further train it on Pitts30k-train for Pitts30k. For fine-tuning, the DHE network and the last 2 encoder layers in backbone are updatable. The model for Pitts30k is fine-tuned on Pitts30k-train for 40 epochs (5k iterations per epoch), while the model for others is fine-tuned on MSLS-train for 2 epochs (10k iterations per epoch). We use 2 hard negative images in a triplet.

\subsection{Ablation Study}
We conduct several ablation experiments on the Pitts30k and MSLS (val) datasets to validate the design of our DHE network and training strategy. We demonstrate the effectiveness by comparing performance before and after using the DHE network for re-ranking, as well as before and after using our proposed training strategy for fine-tuning. We also use the solution with RANSAC-based re-ranking as a reference.
\begin{itemize}
	\item\textbf {GeM:} Direct retrieval with the GeM global feature.%, whose model is trained solely on the backbone.
	\item\textbf {GeM+DHE:} GeM feature is used to retrieve candidates, and the homography estimated by DHE network is used to check spatial consistency for re-ranking. The backbone and DHE network are trained independently.
	\item\textbf {GeM+RANSAC:} GeM and RANSAC are used for candidates retrieving and re-ranking, respectively.
	\item\textbf {GeM*:} Direct retrieval with the GeM feature in our DHE-VPR model. The backbone and DHE network are jointly fine-tuned with our training strategy.
	\item\textbf {GeM+DHE*:} GeM and DHE are used for candidates retrieving and re-ranking, respectively. The backbone and DHE network are jointly fine-tuned with the proposed training strategy. i.e. our complete method.
\end{itemize}

\begin{table}[t!]
  \setlength{\tabcolsep}{1.mm}{
  \renewcommand{\arraystretch}{0.45}
\centering
\begin{tabular}{cccccccc}
\toprule
\multirow{2}{*}{Method} & \multicolumn{3}{c}{Pitts30k} & & \multicolumn{3}{c}{MSLS val} \\ \cmidrule{2-4} \cmidrule{6-8}
  & R@1  & R@5  & R@10  & & R@1  & R@5  & R@10  \\ \cmidrule{1-8}
GeM & 83.1 & 92.8 & 95.2 & & 80.5 & 90.4 & 92.3 \\
GeM+DHE   & 87.8 & 94.3 & 95.9 & & 81.2 & 90.1 & 92.8 \\ 
GeM+ransac   & 88.3 & 94.4 & 95.8 & & 81.8 & \textbf{91.6} & \textbf{93.1} \\
GeM* & 83.9 & 93.3 & 95.4 & & 80.3 & 90.0 & 92.3 \\
GeM+DHE*  & \textbf{89.4} & \textbf{95.1} &  \textbf{96.2} & & \textbf{84.1} &  91.4 &  \textbf{93.1} \\ \bottomrule
\end{tabular}}
\vspace{-0.1cm}
\caption{Ablations on geometric check and training strategy.}
\vspace{-0.45cm}
\label{tab:ablation}
\end{table}

\begin{figure}[!t]
	\centering
	\includegraphics[width=0.9\linewidth]{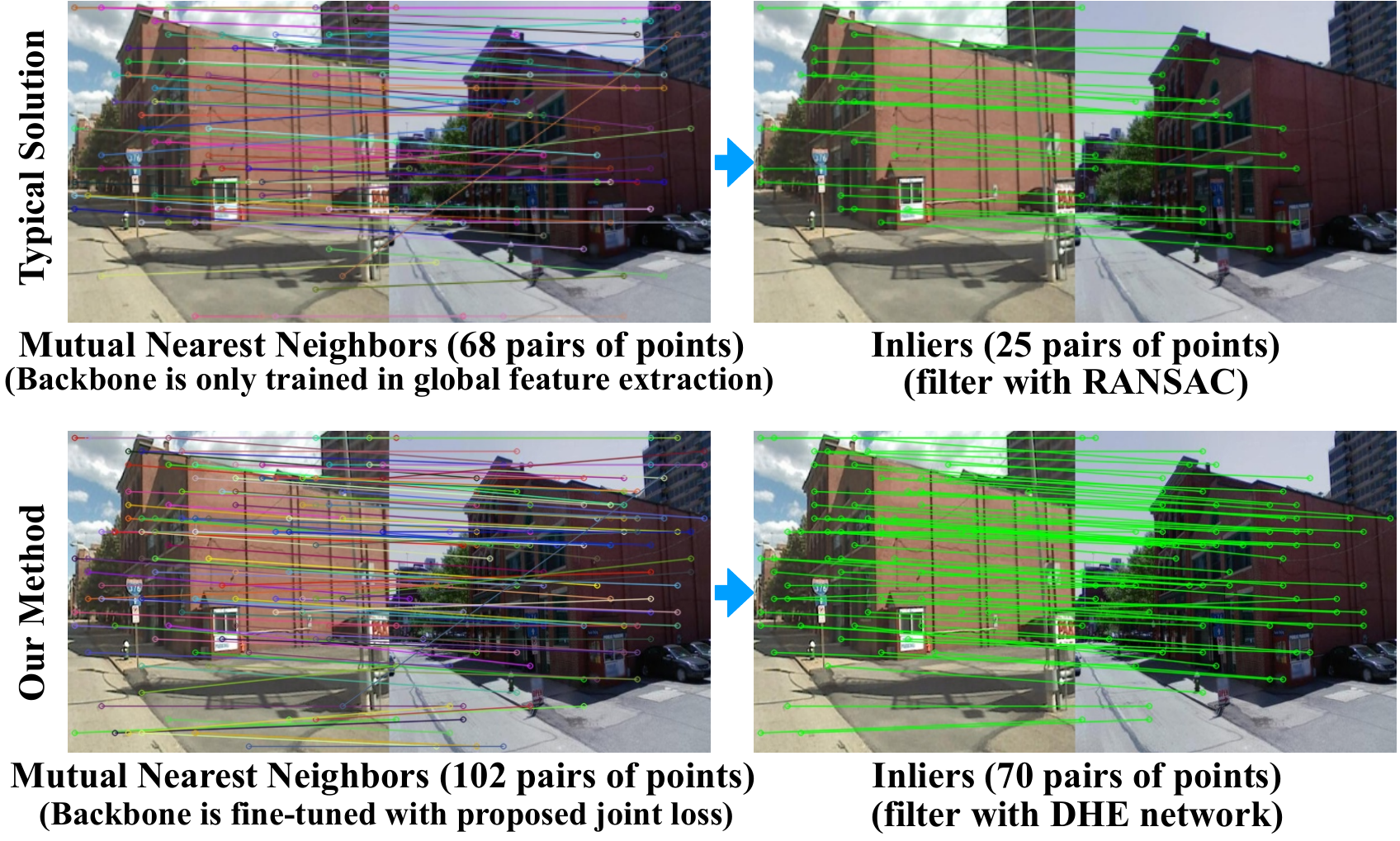}
    \vspace{-0.15cm}
	\caption{Qualitative comparison of typical solution and our method. The top is the typical solution using RANSAC, which is non-differentiable, i.e. the backbone is only trained in global feature extraction. The bottom is ours, which yields more mutual nearest neighbors and inliers than the top. (When the re-projection threshold $\theta$ is set to 1.5× the patch size, ours yields 39 inliers, which is also more than the top.)}
 \label{fig:ablationfig}
 \vspace{-0.5cm}
\end{figure}

Table \ref{tab:ablation} displays the results of different ablated versions. Before fine-tuning with our training strategy (i.e., GeM, GeM+DHE, GeM+RANSAC), either the solution of using DHE network or RANSAC to check geometric consistency for re-ranking improves the performance on the one-stage retrieval (GeM). This indicates that it is feasible for us to use the DHE network to fit homography for geometric verification. After fine-tuning with our proposed training strategy (i.e., GeM* and GeM+DHE*), there is no significant difference in performance between GeM* and GeM, while GeM+DHE* outperforms GeM+DHE. The former shows that fine-tuning with the joint loss does not affect the performance of global features. Meanwhile, the latter indicates that the joint fine-tuning strategy enables the backbone to output better local features (i.e. the features are more suitable for local matching). GeM+DHE* surpasses GeM+RANSAC, demonstrating the superiority of differentiable deep homography over non-differentiable RANSAC. That is, thanks to the differentiable DHE process, the error can assist the backbone in parameter updating via back-propagation. This leads to that although our proposed REI loss uses the number of inliers yielded by RANSAC as supervision, our method outperforms the typical method directly using RANSAC. Fig. \ref{fig:ablationfig} depicts this visually. The (correct) mutual nearest neighbors of our method are significantly more than that of the typical solution, resulting in a larger number of inliers of our method than the typical solution after geometric verification.

\subsection{Different Strategies with Deep Homography for Re-ranking.}
In VPR, there are at least two options for re-ranking using deep homography. One is to regress the homography to align two images taken at a same place with different viewpoints, as in GeoWarp. The other is to fit the homography for geometric verification, which is what we do. We demonstrate the benefit of using deep homography for geometric verification by comparing their performance.

To be fair and not affect the performance of GeoWarp, we replace our backbone with VGG16 \cite{vgg}, which is used in GeoWarp. The models tested on Pitts30k and MSLS were trained on their corresponding training sets. Since the input of the Homography Regression Module in our DHE network is similarity maps (as in GeoWarp), which is more general than feature maps for different feature distributions, we can directly use the DHE network trained for CCT-14 backbone without any re-training. Furthermore, we do not fine-tune the model with our proposed training strategy to ensure that the parameters of the backbone in GeoWarp and our model are identical. In this ablation study, only top-5 candidates are re-ranking and the evaluation metric is Recall@1 under three error tolerances: 10m, 25m, and 50m (to be consistent with the GeoWarp study \cite{geowarp}).

\begin{table}[!t]
\setlength{\tabcolsep}{0.8mm}{
\centering
\begin{tabular}{cccccccccc}
\toprule
\multirow{2}{*}{Method} & Runtime & \multicolumn{3}{c}{Pitts30k} & & \multicolumn{3}{c}{MSLS val} \\ \cline{3-5} \cline{7-9}
& (s)                  & 10m  & 25m  & 50m  & & 10m  & 25m  & 50m  \\ \hline
VGG16+GeM & -                           & 55.4 & 70.6 & 76.3 & & 44.6 & 72.3 & 78.1 \\
GeoWarp & 0.201  & 65.3 & 79.2 & 83.1 & & 48.0 & 74.2 & 79.3 \\ 
Ours-VGG16 & \textbf{0.026} & \textbf{67.6} & \textbf{80.4} & \textbf{84.0} & & \textbf{51.9} & \textbf{77.8} & \textbf{81.2} \\ \bottomrule
\end{tabular}}
\vspace{-0.1cm}
\caption{Comparison of different strategies with deep homography for re-ranking. ``Runtime" is re-ranking latency. We calculated R@1 under three different thresholds.}
\label{tab:withGeoWarp}
\vspace{-0.2cm}
\end{table}

The results are shown in Table \ref{tab:withGeoWarp}. VGG16+GeM is the baseline without re-ranking. On the Pitts30k dataset, both GeoWarp and our method can significantly improve the performance with re-ranking. This suggests that, in urban scenes, it is a good choice whether use deep homography to align images or to check the geometric consistency (although our method still has some advantages over GeoWarp). However, on the MSLS dataset with some suburban scenes, our method significantly outperforms GeoWarp. Our method achieves an absolute improvement on Recall@1 of 5.5\% over GeM within 25m tolerance, which is nearly three times the 1.9\% increase provided by GeoWarp. This supports the theory that it is difficult to align two images presenting different planes using homography in the suburban scene. And deep homography for geometric verification is a more robust solution. Besides, the re-ranking runtime of a single query on Pitts30k is provided in Table \ref{tab:withGeoWarp}. Since GeoWarp requires feature extraction and matching on the aligned image pairs again after warping, our method does not. It gives our method a significant advantage in efficiency.

\subsection{Comparisons with SOTA Methods}
\begin{table*}
  \centering
  \small
  \setlength{\tabcolsep}{0.65mm}{
  \begin{tabular}{@{}l|ccc|ccc|ccc|ccc|ccc|ccc@{}}
  \toprule
\multirow{2}{*}{Method}   & \multicolumn{3}{c|}{Pitts30k} & \multicolumn{3}{c|}{MSLS val} & \multicolumn{3}{c|}{MSLS challenge} & \multicolumn{3}{c|}{Nordland test} & \multicolumn{3}{c|}{St. Lucia} & \multicolumn{3}{c}{Average}\\
\cline{2-19}
& R@1 & R@5 & R@10 & R@1 & R@5 & R@10 & R@1 & R@5 & R@10 & R@1 & R@5 & R@10 & R@1 & R@5 & R@10 & R@1 & R@5 & R@10 \\
\hline
NetVLAD & 84.0 &92.8 & 94.9 & 52.4 &64.7 & 69.4 &31.5 & 42.1 & 46.2 & 10.9 & 19.2  & 24.5 & 49.0 & 68.1 & 76.2 & 45.6 & 57.4 & 62.2\\
SFRS & \textbf{89.4} & 94.7 & 95.9 & 69.2 & 80.3 & 83.1 & 41.6 & 52.0 &56.3  & 22.0 & 35.0 & 41.8 & 75.4 & 86.2 & 90.2 & 59.5 & 69.6 & 73.5\\
CosPlace & 88.4 & 94.5 & 95.7 & 82.8 & 89.7 & 92.0  & 61.4 & 72.0 & 76.6 & \underline{64.1} & \underline{83.2} & \textbf{89.2} &94.9 &  97.5 &  98.7 &78.3 &  \underline{87.4} &  \underline{90.4}\\
GeM*  & 83.9 & 93.3 & 95.4 & 80.3 & 90.0 & 92.3 &57.1 &\underline{74.7} &\underline{80.1}  &46.8 & 72.5 & 81.3 &96.0  &98.8  & \underline{99.5} & 72.8 & 85.9 & 89.7\\
\hline
SP-SuperGlue & 87.2 & 94.8 & \textbf{96.4} & 78.1 & 81.9 & 84.3 & 50.6 &56.9  &58.3  &  25.8  & 35.4  & 38.2  & 86.5 & 92.1 & 93.4 & 65.6 & 72.2 & 74.1  \\
Patch-NetVLAD-s & 87.5 & 94.5 & 96.0 & 77.8 & 84.3 & 86.5 &48.1 &59.4  & 62.3  & 44.2 & 57.5 & 62.7 & 90.2 & 93.6 & 95.0 & 69.6 & 77.9 & 80.5\\
Patch-NetVLAD-p & 88.7 & 94.5 & 95.9  & 79.5 & 86.2 & 87.7 & 48.1 &57.6 & 60.5 & 51.6 & 60.1 & 62.8 & 93.9 & 95.5 & 96.2 & 72.4 & 78.8 & 80.6\\
TransVPR &  \underline{89.0} &  \underline{94.9} &  \underline{96.2} & \textbf{86.8} & \underline{91.2} & \underline{92.4} & \textbf{63.9} & 74.0  & 77.5  &  61.3 &  71.7 &  75.6 &  \underline{98.7} & \underline{99.0} & 99.2 & \underline{79.9} & 86.2 & 88.2\\
ETR-D & 84.2 & 91.6 & 93.8 & 79.3 & 88.0 & 89.6 & 50.6 &62.1 & 65.8 & - & - & - & - & - & - & - & - & -\\
\hline
DHE-VPR (ours) & \textbf{89.4} & \textbf{95.1} &  \underline{96.2} &  \underline{84.1} &  \textbf{91.4} &  \textbf{93.1} &  \underline{61.7} &  \textbf{78.2} &  \textbf{82.6} & \textbf{67.4} & \textbf{85.4} & \underline{88.9} & \textbf{99.1} & \textbf{99.6} & \textbf{99.7} & \textbf{80.3} & \textbf{89.9} & \textbf{92.1}\\
\bottomrule
\end{tabular}}
\vspace{-0.1cm}
\caption{Comparison to SOTA methods on benchmark datasets. The best is highlighted in \textbf{bold} and the second is \underline{underlined}.}
\vspace{-0.3cm}
\label{tab:compare_SOTA}
\end{table*}

\begin{figure*}[!t]
	\centering
	\includegraphics[width=0.87\linewidth]{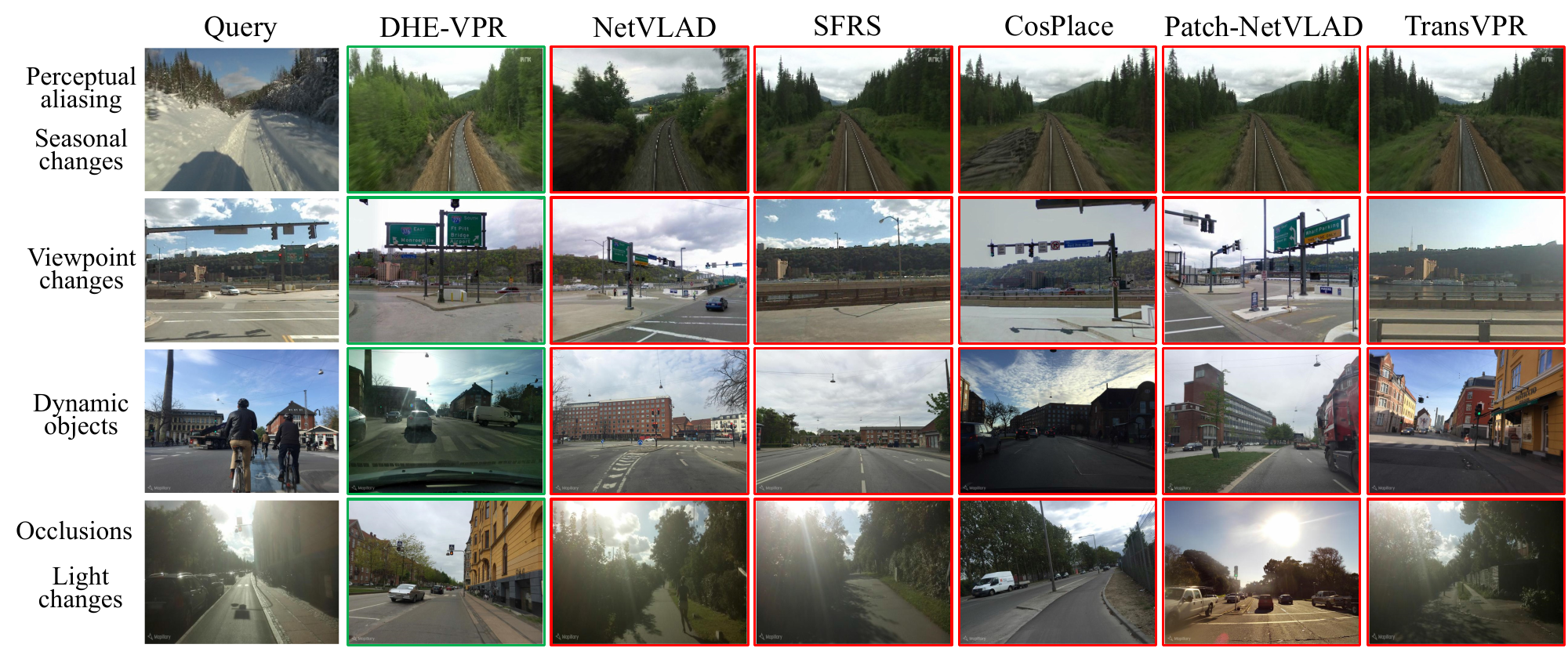}
    \vspace{-0.1cm}
	\caption{
		Qualitative results. In these challenging examples, our DHE-VPR successfully retrieves the correct images, while all other methods yield false places. In the second example, some other methods actually get database images that are geographically close to the query image, but their radius exceeds the threshold (25m). In the last example, the buildings on the left of the query image are occluded by vehicles and vegetation.
	}
 \vspace{-0.4cm}
	\label{fig:result}
\end{figure*}

We compared our DHE-VPR method against several SOTA VPR algorithms, including three one-stage VPR methods using global features for direct retrieval: NetVLAD \cite{netvlad}, SFRS \cite{sfrs} and CosPlace \cite{cosplace}, as well as four two-stage VPR methods with re-ranking: SP-SuperGlue \cite{sp,sg}, Patch-NetVLAD \cite{patchvlad}, TransVPR \cite{transvpr} and ETR-D \cite{etr}. We also show the direct retrieval result using GeM in our pipeline (denoted as GeM*).

Two Patch-NetVLAD versions are used in our experiments, i.e., Patch-NetVLAD-s (speed-focused) and Patch-NetVLAD-p (performance-focused). SP-SuperGlue is first used for VPR in Patch-NetVLAD work, in which NetVLAD is used to retrieve candidates, and the SuperGlue \cite{sp} matcher is applied to match the SuperPoint \cite{sg} key-point features for re-ranking. TransVPR and ETR-D employ the transformer model (same as ours). Patch-NetVLAD-p and TransVPR fit homography for geometric check, but they use RANSAC, whereas we use deep homography. Besides, they use the multi-scale fusion \cite{patchvlad} or multi-level aggregation \cite{transvpr} on features to boost performance, but we do not.

The quantitative results of our DHE-VPR compared with other methods are shown in Table \ref{tab:compare_SOTA}. On all datasets, our DHE-VPR achieves the best Recall@5. And the average performance (R@1, R@5, and R@10) of our method on all datasets is also the best among all methods. CosPlace is the SOTA one-stage VPR method because it was trained on the very large-scale SF-XL dataset \cite{cosplace}, whereas our method was not trained on such a dataset. Although our model has no advantage when using only global features for retrieval, our complete DHE-VPR outperforms CosPlace thanks to geometric verification for re-ranking. Especially on Nordland, which is prone to perceptual aliasing, our method has a 20.6\% absolute increase on R@1 than without re-ranking. TransVPR is an excellent two-stage VPR method, with the best R@1 on MSLS val and MSLS challenge. However, our method outperforms it in most results, achieving absolute improvements of 4.2\% and 13.7\% on R@5 on the MSLS challenge and Nordland, respectively. Besides, all methods use 640$\times$480 resolution images except our method, which uses the 384$\times$384 resolution. This enables our method to perform well on the low-resolution Nordland dataset as well. The qualitative results in Fig. \ref{fig:result} illustrate our method is highly robust against condition (e.g. light, season) and viewpoint changes, and less susceptible to perceptual aliasing than other methods. 

\subsection{Runtime Analysis}

\begin{table}
  \centering
  \small
\begin{tabular}{lccc}
\toprule
Method & \begin{tabular}[c]{@{}c@{}}Extraction\\ Time (s)\end{tabular} &         \begin{tabular}[c]{@{}c@{}}Matching\\ Time (s)\end{tabular} & \begin{tabular}[c]{@{}c@{}}Total \\ Time (s)\end{tabular} \\
\hline
SP-SuperGlue & 0.042 & 6.639 & 6.681 \\
Patch-NetVLAD-s & 0.186 & 0.551 & 0.737 \\
Patch-NetVLAD-p & 0.412 & 10.732 & 11.144 \\
TransVPR & 0.008 & 3.010 & 3.018 \\
\hline
DHE-VPR (Re100) & 0.006 & 0.264 & 0.270 \\
DHE-VPR (ours) & \textbf{0.006} & \textbf{0.098} & \textbf{0.104} \\
\bottomrule
\end{tabular}
\vspace{-0.1cm}
\caption{Runtime (feature extraction time and matching time) of two-stage methods on Pitts30k. DHE-VPR(Re100) uses the proposed architecture to re-rank top-100 candidates.}
\label{tab:time}
\vspace{-0.2cm}
\end{table}

\begin{figure}[!t]  
	\centering
	\includegraphics[width=0.62\linewidth]{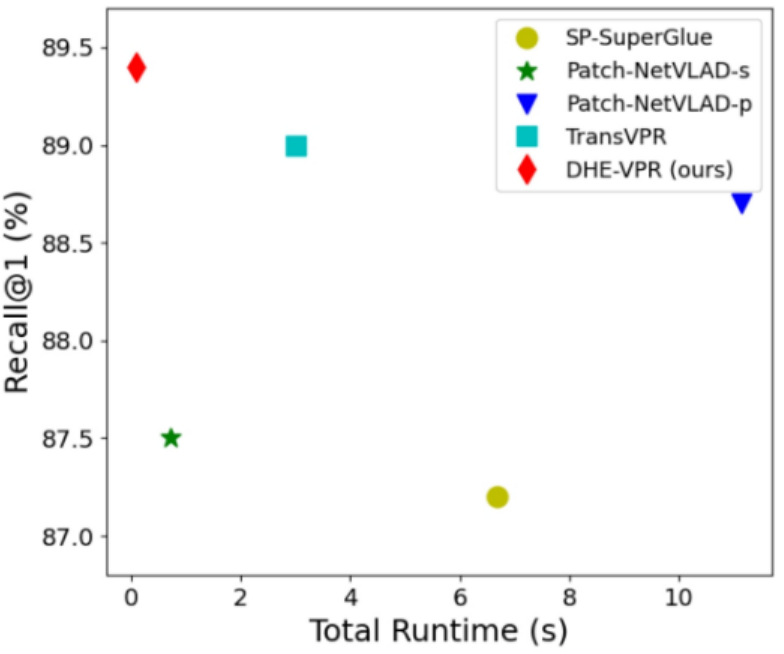}
 \vspace{-0.2cm}
	\caption{Recall@1-Runtime comparison of two-stage VPR methods on the Pitts30k dataset.}
 \label{fig:recalltime}
 \vspace{-0.4cm}
\end{figure}

We evaluate the computational efficiency of the VPR systems using the runtime (feature extraction time and matching time) of a single query on the Pitts30k test dataset. We compare the proposed DHE-VPR with the other two-stage VPR methods. The results are shown in Table \ref{tab:time}. Since we use the compact backbone (CCT) and the input image is only 384$\times$384 resolution, our method has an advantage in feature extraction runtime. And the matching time of our method is only 0.098s, which is less than 1/30 of TransVPR and Patch-NetVLAD-p. These two methods use RANSAC for geometric verification. Compared with SP-SuperGlue, which also uses neural networks to match local features (but not based on homography), our method takes less than 1/60 of its matching time. Although Patch-NetVLAD-s uses a fast verification algorithm (Rapid Spatial Scoring), it is still more time-consuming than ours. Considering that our method only re-ranks the top-32 candidates while others re-rank top-100, we also provide the runtime of ours to re-rank top-100 candidates for a more fair comparison, denoted as DHE-VPR(Re100). Its matching runtime and total runtime are still one order of magnitude lower than the RANSAC-based methods (TransVPR and Patch-NetVLAD-p). Fig. \ref{fig:recalltime} simultaneously shows total runtime and R@1. Our method has obvious advantages in both performance and efficiency.

\section{Conclusions}
In this paper, we presented a novel hierarchical VPR architecture, which uses a DHE network to regress homography to check the geometric consistency in the re-ranking stage. It can break through the time-consuming and non-differentiable limitations of the RANSAC algorithm. Meanwhile, we proposed the REI loss to train the DHE network, which can be jointly optimized with the backbone thus making the learned feature maps more suitable for local matching in re-ranking. The experimental results showed that our architecture can outperform several SOTA methods on VPR benchmark datasets. And the runtime of ours is more than one order of magnitude lower than that of the existing two-stage methods using RANSAC for geometric verification. 

\section{Acknowledgments}
This work was supported by the National Key R\&D Program of China (2022YFB4701400/4701402), SSTIC Grant (JCYJ20190809172201639, WDZC20200820200655001), Shenzhen Key Laboratory (ZDSYS20210623092001004), the Project of Peng Cheng Laboratory
(PCL2023A08), and Beijing Key Lab of Networked Multimedia.

\bibliography{aaai24}

\end{document}